%% file: sample.tex
\documentclass[10pt]{article}

\input{hicss51-packages.tex}

\usepackage{xurl}
\title{Kernel-Segregated Transpose Convolution Operation}

 \author{Vijay Srinivas Tida \\
   University of Louisiana at Lafayette \\
   {\underline{vijaysrinivas.tida1@louisiana.edu}} \\\And
   \hspace{5cm} Sai Venkatesh Chilukoti\\
     \hspace{5cm}   University of Louisiana at Lafayette \\
   \hspace{5cm}  {\underline{saivenkatesh.chilukoti1@louisiana.edu} } \vspace{0.2cm}\\
   \hspace{-10cm} Sonya Hsu \\
   \hspace{-10cm}  University of Louisiana at Lafayette \\
  \hspace{-10cm} {\underline{hsiu-yueh.hsu@louisiana.edu}} 
   \\\And
   \vspace{1cm} \\\hspace{-6cm} 
   Xiali Hei\\
   \hspace{-6cm} University of Louisiana at Lafayette \\
   {\hspace{-6cm} \underline{xiali.hei@louisiana.edu}}
   }
  
\begin{document}
\maketitle
\begin{abstract}
Deep learning models having transpose convolution layers requires optimization to deploy in the resource constraint Internet of Things (IoT) devices. The main reason is the presence of zeros at predefined positions in the input feature maps after upsampling layer leads to higher memory and computation load requirements for transpose convolution operations. We propose an algorithmic-level optimization technique based on kernel segregation mechanisms for efficient transpose convolution implementation to address these issues without needing an upsampling layer. Experimental results showed that the proposed approach showed an average of $3.7\times (3.4 \times)$ faster computation using an Intel Xeon CPU (RTX 2070 GPU) than the conventional method. Further, we analyzed the performance using different transpose convolution layers from the popular Generative Adversarial Network (GAN) models and a simple deep learning model with one transpose convolution layer. There is a significant improvement in computation speed and substantial memory savings from the obtained results. 
\end{abstract}

\section{Introduction}

\begin{figure}[htbp]
     \centering
     \begin{small}
     \includegraphics[width =
     \columnwidth]{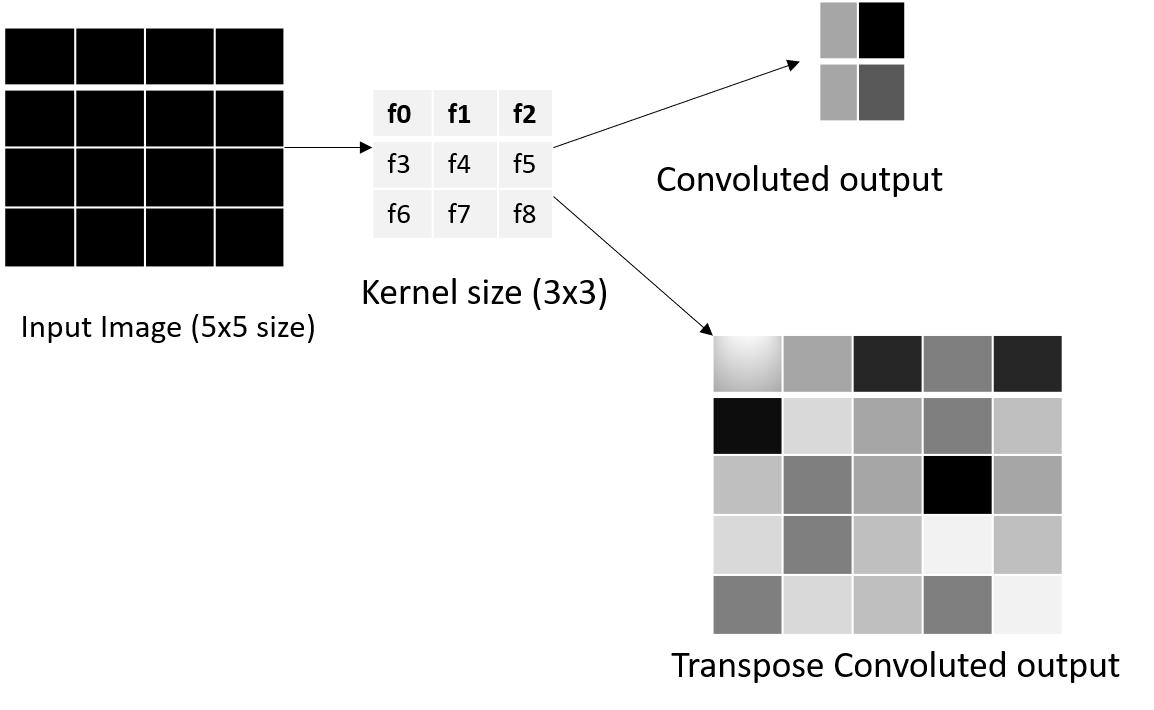}
   \caption{\textbf{Conventional convolution and transpose convolution operations.}   \vspace{-3mm}}
  
    \label{fig1}
    \end{small}
\end{figure}

GANs consist of two parts, namely, the generator and the discriminator. The transpose convolution layer is mainly used in the generator part, whereas the convolution layer is used in the discriminator part. The general overview of the convolution and transpose convolution is illustrated in Fig. \ref{fig1}. Applying the convolution operation on the input feature map will compress the output feature map. In contrast, the transpose convolution operation will expand the output feature map. The output feature map values will be obtained based on selecting the version of the transpose convolution. The transpose convolution with stride one will not be helpful in deep learning applications because of the checkerboard pattern \cite{sharonzhou}. This problem arises due to more values accumulating at the center pixels. Therefore, the transpose convolution layer with a combination of upsampling and convolution layers is used to avoid the checkerboard problem.

\begin{figure}[htbp]
     \centering
     \begin{small}
     \includegraphics[width =  \columnwidth]{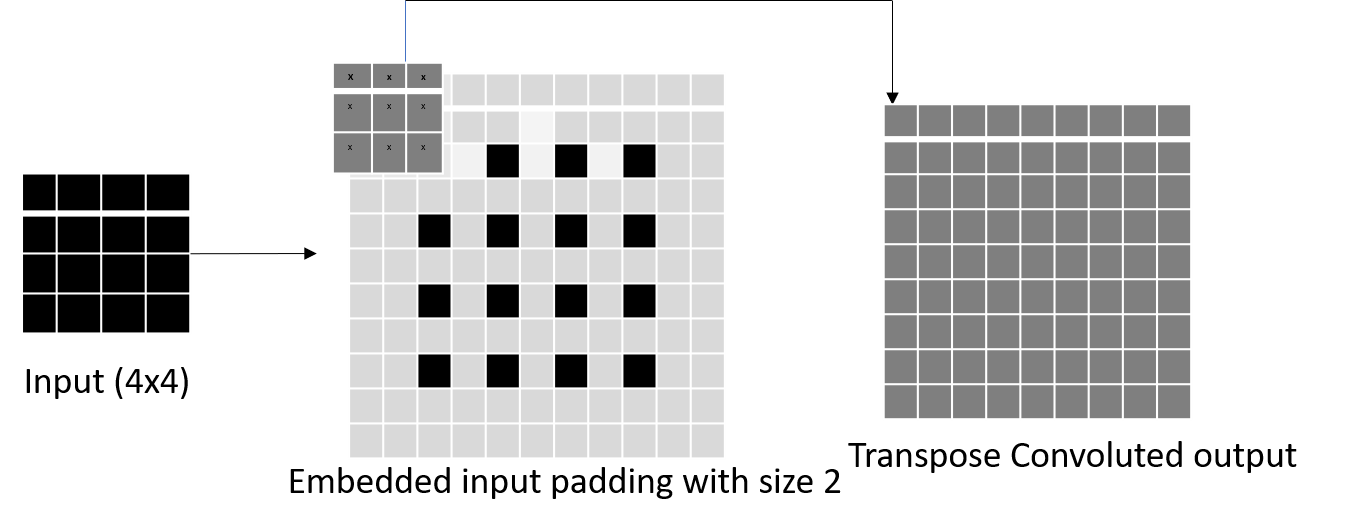}
   \caption{\textbf{Transpose convolution operation with padding factor of 2. \vspace{-3mm}}}
 
    \label{fig2}
      \end{small}
\end{figure}

The transpose convolution layer implementation used in this paper is similar to the research of \cite{yazdanbakhsh2018ganax} and is the standard version used in many popular GANs. The upsampling layer transforms the input feature of size $N \times N$ by embedding zeros after each row and column. The transformation results in  input feature map size to $(2N-1)\times(2N-1)$ after the upsampling process. Applying the convolution operation with a kernel size of $n\times n$  with stride one on the obtained feature map leads to an output feature map of size $(2N-n)\times(2N-n)$. Fig. \ref{fig2} explains the basic transpose convolution operation with the input feature map of size $4\times4$ and a kernel size of $3\times3$. Unnecessary zeros obtained from the upsampled feature map in transpose convolution operation result in excessive data transfers, memory bottlenecks, and wastage of computing resources.

Prior research primarily focused on optimizing convolution and transpose convolution operations using hardware approaches \cite{Marat_dukhan,yan2018gna,yazdanbakhsh2018ganax,yazdanbakhsh2018flexigan,chang2018energy,van2015blis}. These implementations require extra hardware, and some need upsampling layers for efficient transpose convolution implementation. To the best of our knowledge, we are the first to introduce the optimization algorithm for transpose convolution without using an upsampling layer.

% Several upsampling methods under non-learning upsampling and learning-based upsampling techniques were proposed  \cite{wang2020deep}. Non-learnable upsampling techniques like Nearest Neighbors, Bi-Linear Interpolation, Bicubic Interpolation, Bed of Nails, and max-unpooling methods are predefined and invariant on the data. These techniques are task-specific, meaning they do not learn any information from the input data. Interpolation techniques pose additional problems like computational complexity, blurring results, and noise amplification. To overcome these problems, learnable upsampling techniques such as transpose convolution layer or deconvolution layer  \cite{shi2016deconvolution}, sub-pixel layer  \cite{shi2016real} and meta upscale module  \cite{hu2019meta} were proposed. These techniques learn information from the given input data using learnable parameters. Among the learnable upsampling techniques, transpose convolution has become the most popular scheme because of its usage in generative adversarial networks (GANs)  \cite{goodfellow2014generative}. 

% Using different padding factors can change the input feature map's dimensions, which results in the output feature map's dimensions. 

The significant contributions of this paper are as follows:

a. We propose an optimized transpose convolution algorithm using a kernel segregation mechanism to reduce computation load and memory requirement without requiring specialized hardware.

b. We analyze the speed up in computation time and memory savings of our proposed approach using multiple datasets and the transpose convolution layers from popular GANs. 

c. We also analyzed the delay, area, and power consumption of the proposed model using Synopsys DC
compiler with 45nm and 14nm technologies

d. We investigate the performance of the proposed optimization using a simple deep neural network having one transpose convolution layer. Our experimental results indicate a significant improvement in training time without needing an upsampling layer.

The rest of the paper is organized as follows: Section \ref{sec2} explains the background, and literature review, whereas Section \ref{sec3} explains the kernel segregation mechanism along with the optimized transpose convolution process. Section \ref{sec4} interprets the results and finally, Section \ref{sec6} concludes the paper with future research directions.

\section{Background and Literature Review}\label{sec2}

The formula for the convolution using the 2D input array can be expressed in Equation \ref{eq01} \cite{conv}.

\begin{equation}\label{eq01}
\begin{small}
\begin{array}{l}
out[i,j] = \sum_{u=1}^{n}\sum_{v=1}^{n}in[i+u][j+v]*k[u][v], 
\end{array}
\end{small}
\end{equation}

where the array $out$ represents the output feature map with dimension $(N-n+1) \times (N-n+1)$, the array $in$ represents the input feature map of size $N \times N$ and $k$ represents the kernel of size $n \times n$. The element $out[i,j]$ denotes the output feature map's value at $i^{th}$ row and $j^{th}$ column. The variable $in[i+u][j+v]$ represents the input feature map's value at $(i+u)^{th}$ row and $(j+v)^{th}$ column and $k[u][v]$ represents the value of the kernel at $u^{th}$ row and $v^{th}$ column. The same equation is applicable for transpose convolution, but the input dimension will be $(2N-1) \times (2N-1)$ obtained after the upsampling process. 

\subsection{Algorithms for convolution operation} \label{2.1}
The implementation of the general convolution algorithm is clearly explained in  \cite{anderson2020high}. GEMM-based algorithms use computations in the convolution operator as a GEneral Matrix Multiplication with highly optimized Basic Linear Algebra Subprograms (BLAS) \cite{goto2008anatomy, van2015blis}. Many deep learning frameworks, including Tensorflow \cite{abadi2016tensorflow}, PyTorch \cite{paszke2017automatic}, and Caffe \cite{jia2014caffe}, use GEMM-based algorithms introduced in \cite{chellapilla2006high}. However, these algorithms need patch matrices that require more memory storage and bandwidth to perform convolution operations. Several works were proposed to reduce computation costs by reducing the multiplications required for convolution separately for CPU and GPU applications \cite{keshab, chen2018tvm, georganas2018anatomy, heinecke2016libxsmm, zhang2018high}.

Fast convolution algorithms using Fourier or Winograd transformations were introduced \cite{vasilache2014fast,lavin2016fast}. \cite{bhattacharya2016sparsification} used a separable convolution operation by converting 2D kernels into the row and column kernels for mobile and embedded platforms. \cite{wang2019parallel} proposed a parallel convolution algorithm and showed their performance on multi-core CPUs. The performance evaluation shows a factor ranging from 1.0 to $5.17 \times$ than GEMM-based implementation. \cite{anderson2020high} used smaller patches for computing convolution to reduce the memory overhead. Later, an indirect convolution algorithm helped to eliminate expensive and memory-intensive im2col transformations and replace the im2col buffer with a much smaller indirection buffer \cite{Marat_dukhan}.
% The proposed approach has a limitation that cannot apply to the backward propagation of convolution layers. 
Therefore, the general convolution method used for transpose convolution implementation is advantageous as it requires less memory than other optimized algorithms. Still, the computation time will be longer and beneficial for resource constraint devices due to lower memory requirements. Furthermore, this approach will make the backpropagation process also easier for transpose convolution during the training process of neural networks. However, the proposed approach has a significant limitation: It does not support backward propagation for convolutional layers. Moreover, the proposed algorithms might not be efficient for transpose convolution implementation because of nearly 70\% zeros embedded in the upsampled input feature map. 

\subsection{Hardware accelerators for transpose convolution operations}

\cite{yazdanbakhsh2018ganax,yazdanbakhsh2018ganax} designed hardware accelerators using Application Specific Integrated Circuit (ASIC) and Field Programmable Gate Array (FPGA) for implementing transpose convolution efficiently. However, these hardware accelerators avoid unnecessary computations but demand more memory because of the upsampling layer. On the other hand, efficient implementation of transpose convolution was made using systolic arrays by Huynh \textit{et al.} and filed a patent through Amazon Technologies \cite{TCPT}. However, the authors didn't explain the usage of the proposed hardware for the backpropagation process using systolic arrays. Moreover, the proposed methods above require dedicated hardware that might not be easily available to researchers.

%\subsection{Need for optimized transpose convolution algorithm}
%need to draw the conventional and proposed approach
\section{Methodology} \label{sec3}

%This section will first introduce the kernel segregation mechanism. 
% Later in the generalization of the kernel segregation mechanism, we generalize the method that segregating a kernel of size $N \times N$ into four sub-kernels, and the padding effect for the optimized algorithm is explained. After that, the optimization of the transpose convolution operation using segregated sub-kernels is illustrated with the help of an input feature map of size $4 \times 4$. At last, we explain that why our optimized algorithm could be applied to deep learning models with some minor modifications. \XH{duplicated materials, if no space, you can cut off this}

\subsection{Kernel segregation mechanism} \label{3.1}
\begin{figure}[htbp]
     \centering
     \begin{small}
     \includegraphics[width =0.8 \columnwidth]{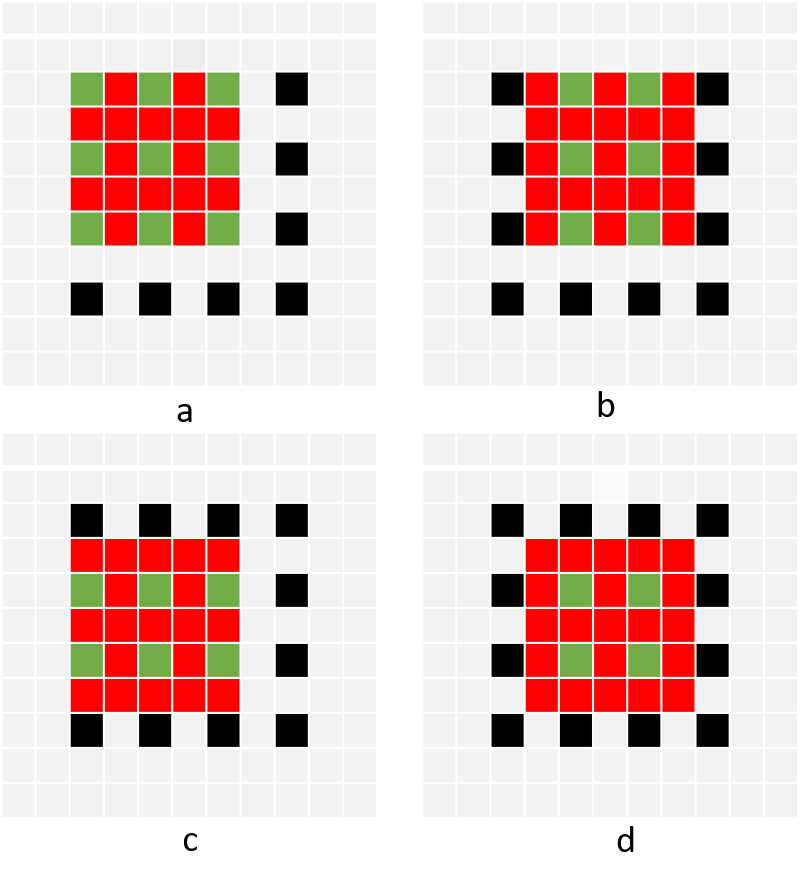}
   \caption{\textbf{Transpose convolution operation. %The red color indicates the zero values in the input, and the green indicates the position where kernel elements are effective. 
   a), b), c), and d) parts show the computation pattern, throughout the input feature map. %There will be four cases in which the kernel can be segregated into four sub-kernels. Here we take the kernel size of $5\times5$ and input a feature map of size $4\times4$ for illustration purposes (Ignore the padding effect). 
   \vspace{-3mm}
   }}
   \label{fig3}
   \end{small}
    
\end{figure}

This process involves segregating the original kernel into four sub-kernels based on the upsampled input feature map pattern. In the input feature map, zeros are usually embedded along each row and column after every element in a predefined manner, as shown in Fig. \ref{fig3} after the upsampling process. Four common cases will arise when the original kernel slides through the input feature map. The red dots indicate that the values are zeros in the corresponding input feature map. The kernel elements are inactive at these positions and need to be discarded. An inactive state means that the multiplication operations will give zero at the related positions. The green dots indicate that the values are non-zero in the corresponding input feature map. The kernel elements that are in the active state should be considered for our segregation mechanism. An active state means that the multiplication operation is effective in these locations.

Note that we assume the indexing of elements starts at (0,0) on the input feature map. In the first case, as in Fig. \ref{fig3}a, only a combination of even row and even column elements from the original kernel are in the active state, and others are inactive. In the second case, as in Fig. \ref{fig3}b, only a combination of even row and odd column element operations from the original kernel is in the active state, and all other element positions are useless. In the third case, as in Fig. \ref{fig3}c, only a combination of odd row and even column elements is used for computation, and others remain unused. Similarly, as in Fig. \ref{fig3}d, only a combination of odd row and odd column elements is necessary for the fourth case, and the others remain unnecessary. We can indirectly perform four convolution operations on the same input feature map if four cases are appropriately analyzed. This significant observation will help design the optimization algorithm using kernel segregation. Finally, there will be some offsets based on the particular activation set. Here we ignored the padding effect and assumed the input elements started from the third row and third column. However, the above process still holds for different padding factors, but the order of cases might change. We will explain these offsets and the padding effect in Section \ref{E}.
%how to refer to sub-part of the diagram

\subsection{Generalization of kernel segregation mechanism} 
\vspace{-1mm}
We can apply the kernel segregation mechanism to any kernel size of $N \times N$ such that N is odd. The general matrix representation of four sub-kernels can be seen in Equations  \ref{m2}, \ref{m3}, \ref{m4}, and \ref{m5}, respectively from original kernel of size $N\times N$. The four sub-kernels $K_{1}, K_{2}, K_{3}, K_{4}$ are formed by accessing the corresponding locations from the original kernel $K$. To obtain the first sub-kernel $K_{1}$, the values along with the alternate columns and alternate rows, which start from (0,0)$^{th}$ element, are accessed from the original kernel K. Similarly, the remaining three sub-kernels $K_{2}, K_{3}, K_{4}$ formed by accessing the elements starting with (0,1)$^{th}$, (1,0)$^{th}$, and (1,1)$^{th}$ elements of the original kernel $K$, respectively. These four sub-kernels will help perform the four convolution operations on the given input feature map based on the data patch taken each time. The final sizes of four sub-kernels will be $\lceil N/2 \rceil \times \lceil N/2 \rceil$, $\lceil N/2 \rceil \times \lfloor N/2 \rfloor$, $\lfloor N/2 \rfloor \times \lceil N/2 \rceil$, and $\lfloor N/2 \rfloor\times \lfloor N/2 \rfloor$, respectively. We use $N_{11} \times N_{12}$, $N_{21} \times N_{22}$, $N_{31} \times N_{32}$, and $N_{41} \times N_{42}$ as sizes for four segregated kernels. Here, $\lceil. \rceil$ represents the ceiling function and $\lfloor. \rfloor$ represents the floor function. However, the arrangement of elements will vary if an even ordered kernel is used and still follows the same process.

\begin{gather}
\begin{small}
 K
 %\begin{bmatrix} \Phi_{11} & \Phi_{12} \\ \Phi_{21} & \Phi_{22} \end{bmatrix}
=\begin{bmatrix}
k_{00}&k_{01}&k_{02}&\cdots&k_{0(N-1)} \\
k_{10}&k_{11}&k_{12}&\cdots&k_{1(N-1)}  \\
k_{20}&k_{21}&k_{22}&\cdots&k_{2(N-1))}\\
k_{30}&k_{31}&k_{32}&\cdots&k_{3(N-1)} \\
\vdots & \vdots &\cdots&\cdots&\vdots\\
\vdots & \vdots &\cdots&\cdots&\vdots \\
\vdots & \vdots &\cdots&\cdots&\vdots\\
k_{(N-1)0}&k_{(N-1)1}&k_{(N-1)2}&\cdots&k_{(N-1)(N-1)} 
\end{bmatrix}
\label{m1}
\end{small}
\end{gather}

\begin{gather}
\begin{small}
 K_{00}
 %\begin{bmatrix} \Phi_{11} & \Phi_{12} \\ \Phi_{21} & \Phi_{22} \end{bmatrix}
=\begin{bmatrix}
k_{00}&k_{02}&\cdots &k_{0(N-1)} \\
k_{20}&k_{22}&\cdots &k_{2(N-1)} \\
\vdots & \vdots & \ddots & \vdots\\
k_{(N-1)0}&k_{(N-1)2}&\cdots &k_{(N-1)(N-1)}
\end{bmatrix}	
\end{small}
\label{m2}
\end{gather}

\begin{gather}
\begin{small}
 K_{01}
 %\begin{bmatrix} \Phi_{11} & \Phi_{12} \\ \Phi_{21} & \Phi_{22} \end{bmatrix}
=\begin{bmatrix}
k_{01}&k_{03}&\cdots &k_{0(N-2)} \\
k_{21}&k_{23}&\cdots &k_{2(N-2)} \\
\vdots & \vdots & \ddots & \vdots\\
k_{(N-1)1}&k_{(N-1)3}&\cdots &k_{(N-1)(N-2)}
\end{bmatrix}
\end{small}
\label{m3}
\end{gather}
\begin{gather}
\begin{small}
 K_{10}
 %\begin{bmatrix} \Phi_{11} & \Phi_{12} \\ \Phi_{21} & \Phi_{22} \end{bmatrix}
=\begin{bmatrix}
k_{10}&k_{12}&\cdots &k_{1(N-1)} \\
k_{30}&k_{32}&\cdots &k_{3(N-1)} \\
\vdots & \vdots & \ddots & \vdots\\
k_{(N-2)0}&k_{(N-2)2}&\cdots &k_{(N-2)(N-1)}
\end{bmatrix}
\end{small}
\label{m4}
\end{gather}

\begin{gather}
\begin{small}
 K_{11}
 %\begin{bmatrix} \Phi_{11} & \Phi_{12} \\ \Phi_{21} & \Phi_{22} \end{bmatrix}
=\begin{bmatrix}
k_{11}&k_{13}&\cdots &k_{1(N-2)} \\
k_{31}&k_{33}&\cdots &k_{3(N-2)} \\
\vdots & \vdots & \ddots & \vdots\\
k_{(N-2)1}&k_{(N-2)3}&\cdots &k_{(N-2)(N-2)}
\end{bmatrix}	
\end{small}
\label{m5}
\end{gather}
\begin{comment}
\begin{gather}
 K_{00}
 %\begin{bmatrix} \Phi_{11} & \Phi_{12} \\ \Phi_{21} & \Phi_{22} \end{bmatrix}
=\begin{bmatrix}
k_{00}&k_{02}&\cdots &k_{0(N-2)} \\
k_{20}&k_{22}&\cdots &k_{2(N-2)} \\
\vdots & \vdots & \ddots & \vdots\\
k_{(N-2)0}&k_{(N-2)2}&\cdots &k_{(N-2)(N-2)}
\end{bmatrix}	
\label{K1}
\end{gather}

\begin{gather}
 K_{01}
 %\begin{bmatrix} \Phi_{11} & \Phi_{12} \\ \Phi_{21} & \Phi_{22} \end{bmatrix}
=\begin{bmatrix}
k_{01}&k_{03}&\cdots &k_{0(N-1)} \\
k_{21}&k_{23}&\cdots &k_{2(N-1)} \\
\vdots & \vdots & \ddots & \vdots\\
k_{(N-2)1}&k_{(N-2)3}&\cdots &k_{(N-2)(N-2)}
\end{bmatrix}
\label{m3}
\end{gather}
\begin{gather}
 K_{10}
 %\begin{bmatrix} \Phi_{11} & \Phi_{12} \\ \Phi_{21} & \Phi_{22} \end{bmatrix}
=\begin{bmatrix}
k_{10}&k_{12}&\cdots &k_{1(N-2)} \\
k_{30}&k_{32}&\cdots &k_{3(N-2)} \\
\vdots & \vdots & \ddots & \vdots\\
k_{(N-1)0}&k_{(N-1)2}&\cdots &k_{(N-1)(N-2)}
\end{bmatrix}
\label{m4}
\end{gather}

\begin{gather}
 K_{11}
 %\begin{bmatrix} \Phi_{11} & \Phi_{12} \\ \Phi_{21} & \Phi_{22} \end{bmatrix}
=\begin{bmatrix}
k_{11}&k_{13}&\cdots &k_{1(N-2)} \\
k_{31}&k_{33}&\cdots &k_{3(N-2)} \\
\vdots & \vdots & \ddots & \vdots\\
k_{(N-1)1}&k_{(N-1)3}&\cdots &k_{(N-1)(N-2)}
\end{bmatrix}	
\label{m5}
\end{gather}
\end{comment}
%\subsection{Padding effect on Transpose Convolution using Segregated Kernels}
\subsection{Optimization of the transpose convolution operation using segregated kernels}\label{E}
\begin{figure}[htbp]
     \centering
     \begin{small}
     \includegraphics[width = \columnwidth]{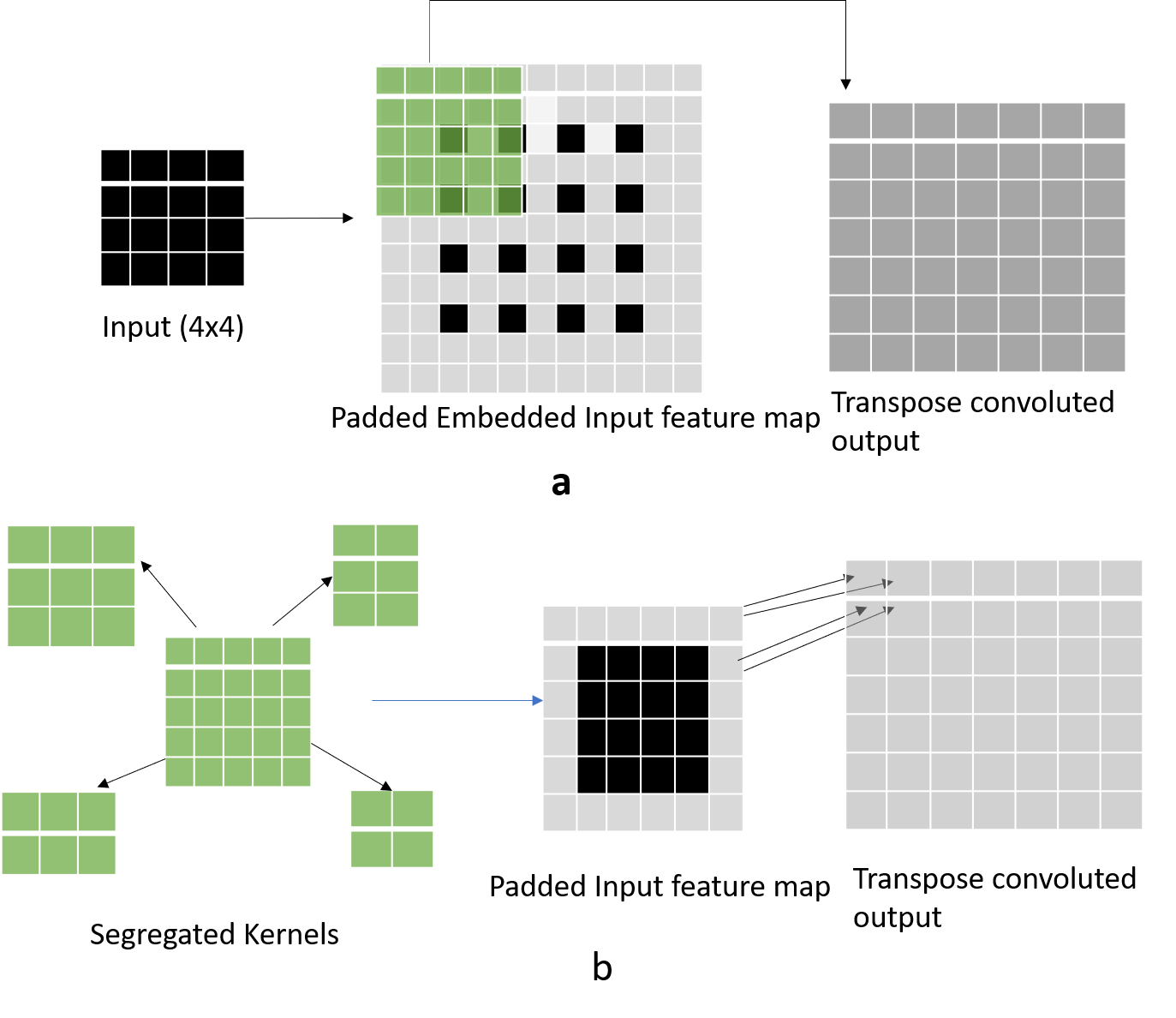}
   \caption{\textbf{Comparison of the a) conventional transpose convolution with the b) proposed optimized technique. }} 
    \label{fig5}
   \end{small}
   
\end{figure}

The conventional transpose convolution and the proposed optimized transpose convolution process can be seen in Fig. \ref{fig5}. Fig \ref{fig5}a represents the input feature map of size $4 \times 4$ embedded with zeros, and a padding factor of 2 is applied. When a kernel of size $5\times5$ slides through the upsampled input feature map, its corresponding output values are obtained sequentially for the conventional method can be observed in Fig. \ref{fig5}a. However, using the proposed kernel segregation mechanism, four output values will be acquired using four sub-kernels can be seen in Fig. \ref{fig5}b. Also, the padding factor for the input feature using four segregated kernels will be different from the original padding factor. For example, if the original padding factor is $P$, the new padding factor will be $\lfloor P/2 \rfloor$. %Additionally, if the original padding factor is odd, we will interchange the new set of four sub-kernels like $K_{4}$, $K_{3}$, $K_{2}$, and $K_{1}$ instead of $K_{1}$, $K_{2}$, $K_{3}$, and $K_{4}$.

\begin{figure}[htbp]
     \centering
     \begin{small}
     \includegraphics[width = \columnwidth]{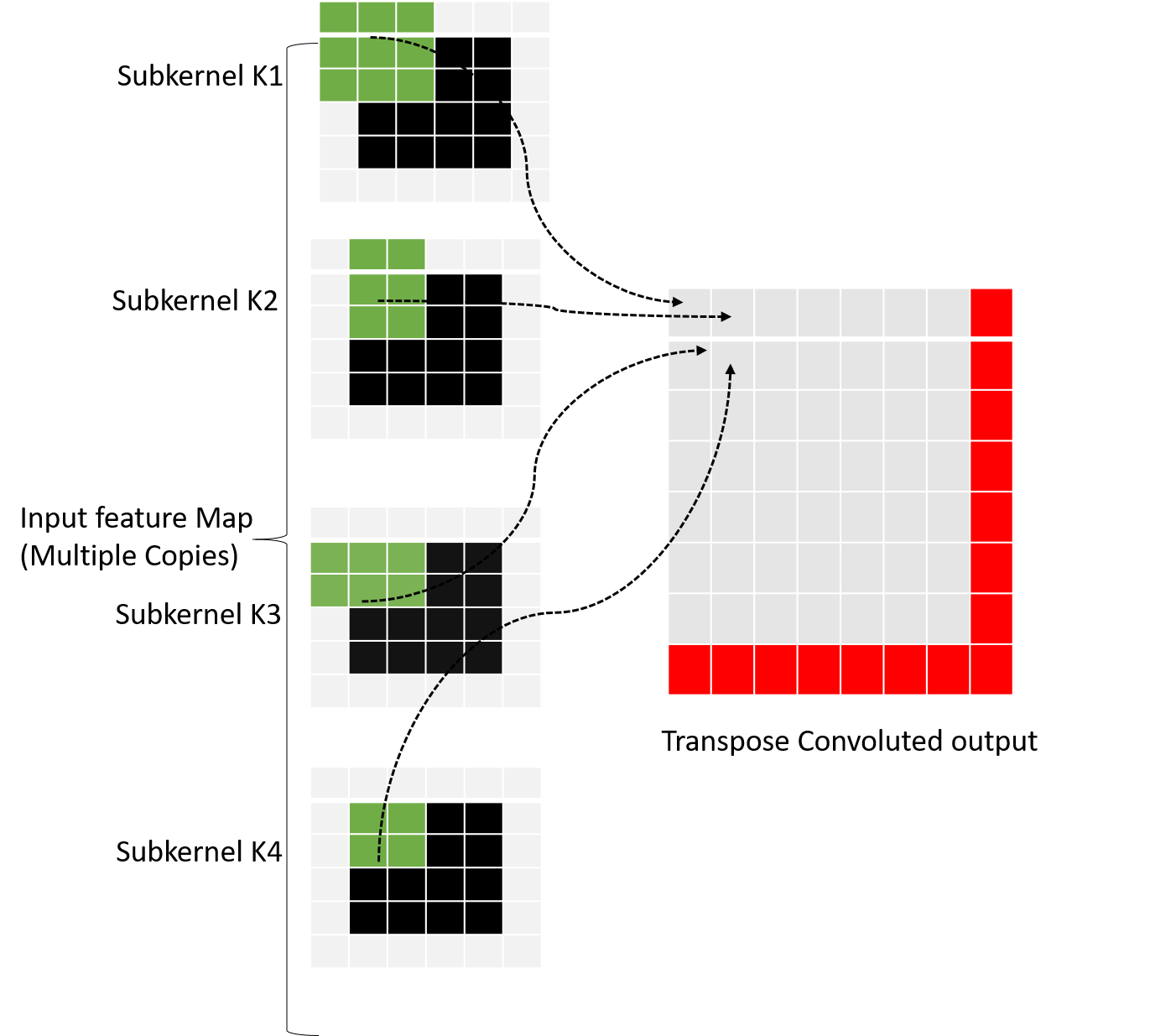}
   \caption{\textbf{Optimized transpose convolution implementation using four sub-kernels with input size $4 \times 4$ and padding with factor 1 (but padding factor is 2 for the original case). %The red lines indicate the unwanted elements need to be checked when optimization is performed.
   }} 
    \label{fig6}
    \end{small}
    \vspace{-3mm}
\end{figure}

Fig. \ref{fig6} illustrates the process of the proposed optimized transpose convolution operation using the kernel segregation mechanism applied on an input feature map of size $4\times4$. Here the padding factor for the input feature map is reduced to 1 from 2 to obtain the exact output feature map from the transpose convolution operation. Next, the convolution operation is applied on the padded input feature map with four sub-kernels to produce four output values at different locations. The first two output locations and the last two output locations are adjacent. On the other hand, one can get the position for the second pair by adding a specific constant from the place of the first pair.

The optimized transpose convolution operation should show four times faster for the ideal case compared to the conventional approach with the same computation load. However, due to the offset problem related to computation in finding specific output locations, there might be some reduction in performance without considering padding and zero embedded time. If the output feature map is of an odd dimension, this continuous process will result in an extra column and row, as indicated in red in Fig. \ref{fig6}. The main reason for the problem is that the optimized algorithm will produce four output feature values in each iteration. The conditional statements can avoid unnecessary computation based on the user requirements. The formulas for calculating the four output feature values by applying the optimization process can be seen in the Equations \ref{eq07}, \ref{eq08}, \ref{eq09}, and \ref{eq10}. 

\begin{equation}\label{eq07}
\begin{small}
\begin{array}{l}

out[2*i][2*j] = \sum_{u=1}^{N_{11}}\sum_{v=1}^{N_{12}}in[i+u][j+v]*K_{1}[u][v], 

\end{array}
\end{small}
\end{equation}
\begin{equation}\label{eq08}
\begin{small}
\begin{array}{l}
out[2*i][2*j+1] = \\\sum_{u=1}^{N_{21}}\sum_{v=1}^{N_{22}}in[i+u][(j+1)+v]*K_{2}[u][v], 
\end{array}
\end{small}
\end{equation}
\begin{equation}\label{eq09}
\begin{small}
\begin{array}{l}
out[2*i+1][2*j] =
\\\sum_{u=
1}^{N_{31}}\sum_{v=1}^{N_{32}}in[(i+1)+u][j+v]*K_{3}[u][v], 
\end{array}
\end{small}
\end{equation}
\begin{equation}\label{eq10}
\begin{small}
\begin{array}{l}
out[2*i+1][2*j+1] =
\\\sum_{u=1}^{N_{41}}\sum_{v=1}^{N_{42}}in[(i+1)+u][(j+1)+v]*K_{4}[u][v], 
\end{array}
\end{small}
\end{equation}

where $out[l][m]$ represents the output feature map located at $l^{th}$ row and $m^{th}$ column; $in[i][j]$ represents the input feature map at the corresponding $i^{th}$ row and $j^{th}$ column; $K_{1}[u][v]$, $K_{2}[u][v]$, $K_{3}[u][v]$ and $K_{4}[u][v]$ represents the sub-kernels $K_{1}$, $K_2$, $K_3$ and $K_4$ obtained after segregation mechanism and their locations at $u^{th}$ row and $v^{th}$ row. The sizes of the corresponding four sub-kernels will be $N_{11} \times N_{12}$, $N_{21} \times N_{22}$, $N_{31} \times N_{32}$, and $N_{41} \times N_{42}$. Here the size of the input feature map will remain the same without upsampled values. The individual output feature map's dimensions depend on the size of the sub-kernels. Finally, the output feature map obtained from the proposed optimization should ensure the same dimensions when conventional transpose convolution is applied. If there are more output values than required, we should discard them.

\begin{figure}[htbp]
     \centering
     \includegraphics[width = \columnwidth]{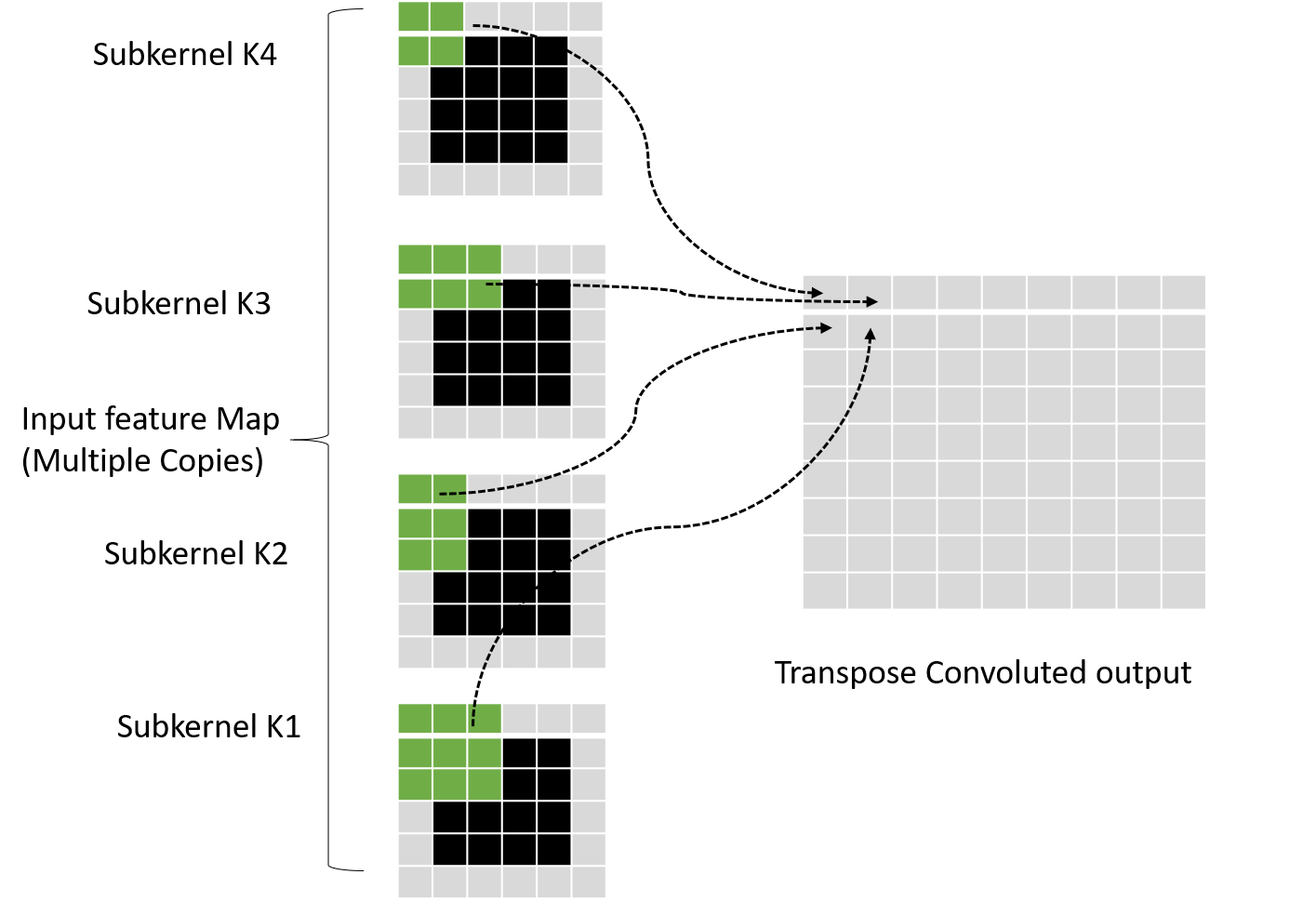}
   \caption{\textbf{Optimized transpose convolution implementation using four sub-kernels with input size $4 \times 4$ and padding with factor 1 (but padding factor is 3 for original case).}}
    \label{fignew_3}
    \vspace{-3mm}
\end{figure}

Fig. \ref{fignew_3} shows the proposed optimization technique when the padding factor is odd, and the kernel size of $5\times5$ is applied on the input feature map. The new padding factor for the input feature map will be one instead of three in the original case to apply the proposed optimization technique. The above exact process and the equations will still hold here, but the order of sub-kernels will change when the four convolution operations are made on the input feature map. The new set of sub-kernels will be $K_{4}$, $K_{3}$, $K_{2}$, and $K_{1}$ instead of $K_{1}$, $K_{2}$, $K_{3}$, and $K_{4}$ for this case. In deep learning applications, the proposed optimization technique can also be used to calculate the kernel and the input gradients during the backward propagation process. Since the proposed approach combines the upsampling and convolution layers, there will be a significant advantage in avoiding unnecessary input gradient computations.

%\subsection{Backpropagation algorithm for the optimized transpose convolution layer }

%Optimized convolution algorithms may not suit the backpropagation process in training a deep learning model. The main reason is that a complicated computation is needed to perform convolution operations. The optimized convolution algorithms are suited only for specific cases based on input feature map size, kernel size, etc. The naive convolution approach can be applied to forward and backward propagation without restrictions for all cases. We used the naive convolution approach with minor modifications in the proposed optimization method. The improvements include accessing the input and output data at predefined locations during the forward propagation process. The same proposed algorithm can be used during backward propagation to calculate the gradients for the kernels and input data. Our approach combines an upsampled layer and a convolutional layer into one layer, avoiding unnecessary input gradients.

\begin{table*}[hbtb]

\centering
\begin{small}
\begin{tabular}{|l|l|l|l|l|l|l|l|l|}
\hline
                                                                           &         &                                                                                   \multicolumn{4}{c|}{Computation time in seconds}                                             &                          &                                                                                                                                                                                          &                                                                     \\ \hline
Data group &\small \begin{tabular}[c]{@{}l@{}}Kernel\\    \\ \end{tabular} &\small \begin{tabular}[c]{@{}l@{}}Conv\\    \\ (GPU)\end{tabular}  &\small \begin{tabular}[c]{@{}l@{}}Conv\\    \\ (CPU)\end{tabular} &\small \begin{tabular}[c]{@{}l@{}}Prop\\    \\ (GPU)\end{tabular} &\small \begin{tabular}[c]{@{}l@{}}Prop\\    \\ (CPU)\end{tabular} &\small \begin{tabular}[c]{@{}l@{}}Speedup\\    \\  (GPU)\end{tabular} &\small \begin{tabular}[c]{@{}l@{}}Speedup\\    \\ (CPU)\end{tabular} &\small \begin{tabular}[c]{@{}l@{}}Memory \\ savings \\ (Bytes)\end{tabular} \\ \hline
Daisy      &\small $5\times5\times3$                                                                    &\small 3.6233     &\small 61.388                                                     &\small 1.018                                                      &\small 15.714                                                     &\small 3.559                                                          &\small 3.906                                                         &\small 1,824,320                                                             \\ \hline
           &\small $4\times4\times3$                                                                    &\small 2.715      &\small 38.978                                                     &\small 0.741                                                      &\small 10.331                                                     &\small 3.663                                                          &\small 3.772                                                         &\small 1,827,900                                                             \\ \hline
           &\small $3\times3\times3$                                                                    &\small 1.7454     &\small 22.491                                                     &\small 0.4916                                                     &\small 6.098                                                      &\small 3.550                                                          &\small 3.6882                                                        &\small 1,824,304                                                             \\ \hline
Dandelion  &\small $5 \times 5 \times 3$                                                                    &\small 5.073      &\small 84.122                                                     &\small 1.4929                                                     &\small 21.496                                                     &\small 3.39                                                           &\small 3.913                                                         &\small 1,824,320                                                             \\ \hline
           &\small $4 \times 4 \times 3$                                                                    & 3.7712     &\small 53.573                                                     &\small 1.043                                                      &\small 14.006                                                     & 3.615                                                          &\small 3.825                                                         &\small 1,827,900                                                             \\ \hline
           &\small $3\times3\times3$                                                                    &\small 2.6008     &\small 30.978                                                     &\small 0.6962                                                     & 8.333                                                      & 3.735                                                          & 3.717                                                         & 1,824,304                                                             \\ \hline
Rose       & $5\times5\times3$                                                                    & 3.5505     & 63.06                                                      & 1.0451                                                     & 15.963                                                     & 3.3972                                                         & 3.9503                                                        & 1824320                                                             \\ \hline
           & $4\times4\times3$                                                                    & 2.736      & 39.945                                                     & 0.7838                                                     & 10.481                                                     & 3.4906                                                         & 3.8112                                                        & 1,827,900                                                             \\ \hline
           & $3\times3\times3$                                                                    & 1.9034     & 23.081                                                     & 0.553                                                      & 6.265                                                      & 3.441                                                          & 3.684                                                         & 1,824,304                                                             \\ \hline
Sunflower  & $5\times5\times3$                                                                    & 3.3974     & 58.809                                                     & 1.0316                                                     & 15.034                                                     & 3.2933                                                         & 3.9117                                                        & 1824320                                                             \\ \hline
           & $4\times4\times3$                                                                    & 2.564      & 37.438                                                     & 0.7326                                                     & 9.829                                                      & 3.4998                                                         & 3.8089                                                        & 1,827,900                                                             \\ \hline
           & $3\times3\times3$                                                                    & 1.6225     & 21.442                                                     & 0.4756                                                     & 5.867                                                      & 3.4114                                                         & 3.6546                                                        & 1,824,304                                                             \\ \hline
Tulip      & $5\times5\times3$                                                                    & 4.5116     & 79.113                                                     & 1.4212                                                     & 20.23                                                      & 3.1749                                                         & 3.9106                                                        & 1,824,320                                                             \\ \hline
           & $4\times4\times3$                                                                    & 3.5148     & 49.918                                                     & 1.0202                                                     & 13.651                                                     & 3.4452                                                         & 3.6567                                                        & 1,827,900                                                             \\ \hline
           & $3\times3\times3$                                                                    & 2.2988     & 28.734                                                     & 0.6851                                                     & 7.963                                                      & 3.3554                                                         & 3.6084                                                        & 1,824,304                                                             \\ \hline
\end{tabular}
\caption{Speedup for GPU and CPU versions and memory savings obtained for Flower dataset for the conventional (Conv) and the Proposed (Prop) approaches}
\label{tab:1}
\end{small}
\end{table*}

\begin{table*}[hbtb]
\centering
\begin{small}

\begin{tabular}{|l|l|l|l|l|l|l|l|}
\hline

                                                                           &                     &                                                                        \multicolumn{4}{c|}{Computation time in seconds}                                                                                                      &                                                                                                                                                             &                                                                     \\ \hline
Dataset                                                                                       & Kernel & \begin{tabular}[c]{@{}l@{}}Conv\\ (GPU)\end{tabular} & \begin{tabular}[c]{@{}l@{}}Conv\\ (CPU)\end{tabular} & \begin{tabular}[c]{@{}l@{}}Prop\\ (GPU)\end{tabular} & \begin{tabular}[c]{@{}l@{}}Prop\\ (CPU)\end{tabular} & \begin{tabular}[c]{@{}l@{}}Speedup\\ (GPU)\end{tabular} & \begin{tabular}[c]{@{}l@{}}Speedup\\ (CPU)\end{tabular} \\ \hline
\multirow{3}{*}{\begin{tabular}[c]{@{}l@{}}MSCOCO \\ 2017\end{tabular}}                       & $5\times 5\times 3$  & 151.49                                               & 951.71                                               & 39.55                                                & 242.252                                              & 3.83                                                    & 3.928                                                   \\ \cline{2-8} 
                                                                                              & $4 \times 4 \times 3$  & 100.22                                               & 618.685                                              & 30.311                                               & 154.301                                              & 3.306                                                   & 4.009                                                   \\ \cline{2-8} 
                                                                                              & $3\times3\times3$  & 60.543                                               & 352.297                                              & 17.973                                               & 92.268                                               & 3.368                                                   & 3.818                                                   \\ \hline
\multirow{3}{*}{\begin{tabular}[c]{@{}l@{}}PASCAL\\ VOC 2012\\ (Classification)\end{tabular}} & $5\times5\times3$  & 210.735                                              & 1395.682                                             & 58.766                                               & 347.362                                              & 3.586                                                   & 4.017                                                   \\ \cline{2-8} 
                                                                                              & $4\times4\times3$  & 144.471                                              & 873.157                                              & 43.636                                               & 226.842                                              & 3.310                                                   & 3.849                                                   \\ \cline{2-8} 
                                                                                              & $3\times3\times3$  & 95.922                                               & 504.292                                              & 29.683                                               & 136.671                                              & 3.232                                                   & 3.689                                                   \\ \hline
\multirow{3}{*}{\begin{tabular}[c]{@{}l@{}}PASCAL \\ VOC 2012\\ (Detection)\end{tabular}}     & $5\times5\times3$  & 35.78                                                & 234.1                                                & 9.658                                                & 57.308                                               & 3.704                                                   & 4.080                                                   \\ \cline{2-8} 
                                                                                              & $4\times4\times3$  & 25.07                                                & 144.72                                               & 6.973                                                & 37.163                                               & 3.595                                                   & 3.894                                                   \\ \cline{2-8} 
                                                                                              & $3\times3\times3$  & 15.709                                               & 90.248                                               & 4.793                                                & 22.293                                               & 3.277                                                   & 4.048                                                   \\ \hline
\end{tabular}

\caption{Speedup for GPU and CPU versions for MSCOCO and PASCAL datasets using conventional (Conv) and proposed (Prop) approaches}
\label{tab:2}
\end{small}

\end{table*}
\section{Results} \label{sec4}
\begin{table*}[hbtb]
\centering
\begin{small}
\begin{tabular}{|l|l|l|l|llll|l|}
\hline
                                                                           &         &            &                                                                       & \multicolumn{4}{c|}{Computation time in seconds}                                                                                                                                                                                                                                                                   &                                                                     \\ \hline
Model                                                                      & \multicolumn{1}{l|}{\begin{tabular}[c]{@{}l@{}}Layer\\ \# \end{tabular}} & Input Size & Kernel Size                                                           & \multicolumn{1}{l|}{\begin{tabular}[c]{@{}l@{}}Conv\\ (GPU)\end{tabular}} & \multicolumn{1}{l|}{\begin{tabular}[c]{@{}l@{}}Prop \\ (GPU)\end{tabular}} & \multicolumn{1}{l|}{\begin{tabular}[c]{@{}l@{}}Conv\\ (CPU)\end{tabular}} & \begin{tabular}[c]{@{}l@{}}Prop \\ (CPU)\end{tabular} & \begin{tabular}[c]{@{}l@{}}Memory \\ savings\\ (bytes)\end{tabular} \\ \hline
\multirow{4}{*}{\begin{tabular}[c]{@{}l@{}} \footnotesize DCGAN/\\ \footnotesize DiscoGAN\end{tabular}} & 2       & $4\times4\times1024$   & $4\times4\times1024\times512$                                                          & \multicolumn{1}{l|}{0.046753}                                                     & \multicolumn{1}{l|}{0.011541}                                                  & \multicolumn{1}{l|}{3.023}                                                        & 0.727                                                     & 495,616                                                             \\ \cline{2-9} 
                                                                           & 3       & $8\times8\times512$    & $4\times4\times512\times256$                                                           & \multicolumn{1}{l|}{0.046085}                                                     & \multicolumn{1}{l|}{0.011381}                                                  & \multicolumn{1}{l|}{3.101}                                                        & 0.6863                                                    & 739,328                                                             \\ \cline{2-9} 
                                                                           & 4       & $16\times16\times256$  & $4\times4\times256\times128$                                                           & \multicolumn{1}{l|}{0.043747}                                                     & \multicolumn{1}{l|}{0.011296}                                                  & \multicolumn{1}{l|}{2.90}                                                         & 0.6598                                                    & 1,254,400                                                           \\ \cline{2-9} 
                                                                           & 5       & $32\times32\times128$  & $4\times4\times128\times3$                                                             & \multicolumn{1}{l|}{0.003049}                                                     & \multicolumn{1}{l|}{0.001551}                                                  & \multicolumn{1}{l|}{0.1363}                                                       & 0.0371                                                    & 2,298,368                                                           \\ \hline
                                                                           &         &            & Total                                                                 & \multicolumn{1}{l|}{0.139639}                                                     & \multicolumn{1}{l|}{0.035769}                                                  & \multicolumn{1}{l|}{9.1603}                                                       & 2.1102                                                    &                                                                     \\ \hline
                                                                           &         &            & \begin{tabular}[c]{@{}l@{}}Total Speedup/\\ Memory saved\end{tabular} & \multicolumn{1}{l|}{}                                                             & \multicolumn{1}{l|}{3.9039}                                                    & \multicolumn{1}{l|}{}                                                             & 4.34                                                      & 4,787,712                                                           \\ \hline
\footnotesize Art-GAN                                                                    & 2       & $4\times4\times512$    & $4\times4\times512\times256$                                                           & \multicolumn{1}{l|}{0.011886}                                                     & \multicolumn{1}{l|}{0.005768}                                                  & \multicolumn{1}{l|}{0.7114}                                                       & 0.1782                                                    & 4,247,808                                                           \\ \hline
                                                                           & 3       & $8\times8\times256$    & $4\times4\times256\times128$                                                           & \multicolumn{1}{l|}{0.011726}                                                     & \multicolumn{1}{l|}{0.002971}                                                  & \multicolumn{1}{l|}{0.7219}                                                       & 0.1652                                                    & 369,664                                                             \\ \hline
                                                                           & 4       & $16\times16\times128$  & $4\times4\times128\times128$                                                             & \multicolumn{1}{l|}{0.021727}                                                     & \multicolumn{1}{l|}{0.00568}                                                   & \multicolumn{1}{l|}{1.3879}                                                       & 0.316                                                     & 627,200                                                             \\ \hline
                                                                           & 6       & $32\times32\times128$  & $4\times4\times128\times3$                                                             & \multicolumn{1}{l|}{0.001582}                                                     & \multicolumn{1}{l|}{0.001532}                                                  & \multicolumn{1}{l|}{0.0359}                                                       & 0.0075                                                    & 67,200                                                              \\ \hline
                                                                           &         &            & Total                                                                 & \multicolumn{1}{l|}{0.046921}                                                     & \multicolumn{1}{l|}{0.015951}                                                  & \multicolumn{1}{l|}{2.8571}                                                       & 0.6669                                                    &                                                                     \\ \hline
                                                                           &         &            & \begin{tabular}[c]{@{}l@{}}Total Speedup/\\ Memory saved\end{tabular} & \multicolumn{1}{l|}{}                                                             & \multicolumn{1}{l|}{2.950}                                                      & \multicolumn{1}{l|}{}                                                             & 4.2841                                                    & 1,871,872                                                           \\ \hline
\footnotesize GP-GAN                                                                     & 2       & $4\times4\times512$    & $4\times4\times512\times256$                                                           & \multicolumn{1}{l|}{0.011847}                                                     & \multicolumn{1}{l|}{0.005787}                                                  & \multicolumn{1}{l|}{0.7114}                                                       & 0.1782                                                    & 247,808                                                             \\ \hline
                                                                           & 3       & $8\times8\times256$    & $4\times4\times256\times128$                                                           & \multicolumn{1}{l|}{0.01171}                                                      & \multicolumn{1}{l|}{0.002952}                                                  & \multicolumn{1}{l|}{0.7219}                                                       & 0.1652                                                    & 369,664                                                             \\ \hline
                                                                           & 4       & $16\times16\times128$  & $4\times4\times128\times64$                                                            & \multicolumn{1}{l|}{0.01167}                                                      & \multicolumn{1}{l|}{0.002895}                                                  & \multicolumn{1}{l|}{0.6995}                                                       & 0.1611                                                    & 627,200                                                             \\ \hline
                                                                           & 5       & $32\times32\times64$   & $4\times4\times64\times3$                                                              & \multicolumn{1}{l|}{0.001574}                                                     & \multicolumn{1}{l|}{0.000852}                                                  & \multicolumn{1}{l|}{0.0659}                                                       & 0.016                                                     & 1,149,184                                                           \\ \hline
                                                                           &         &            & Total                                                                 & \multicolumn{1}{l|}{0.036801}                                                     & \multicolumn{1}{l|}{0.012486}                                                  & \multicolumn{1}{l|}{2.1987}                                                       & 0.5205                                                    &                                                                     \\ \hline
                                                                           &         &            & \begin{tabular}[c]{@{}l@{}}Total Speedup/\\ Memory saved\end{tabular} & \multicolumn{1}{l|}{}                                                             & \multicolumn{1}{l|}{2.9474}                                                    & \multicolumn{1}{l|}{}                                                             & 4.224                                                     & 2,393,856                                                           \\ \hline
\footnotesize EB-GAN                                                                     & 2       & $4\times4\times2048$   & $4\times4\times2048\times1024$                                                         & \multicolumn{1}{l|}{0.188821}                                                     & \multicolumn{1}{l|}{0.046078}                                                  & \multicolumn{1}{c|}{16.0994}                                                      & \multicolumn{1}{c|}{3.598}                                & 991,232                                                             \\ \hline
                                                                           & 3       & $8\times8\times1024$   & $4\times4\times1024\times512$                                                          & \multicolumn{1}{l|}{0.176702}                                                     & \multicolumn{1}{l|}{0.045508}                                                  & \multicolumn{1}{c|}{14.193}                                                       & \multicolumn{1}{c|}{2.919}                                & 1,478,656                                                           \\ \hline
                                                                           & 4       & $16\times16\times512$  & $4\times4\times512\times256$                                                           & \multicolumn{1}{l|}{0.172839}                                                     & \multicolumn{1}{l|}{0.045071}                                                  & \multicolumn{1}{c|}{16.587}                                                       & \multicolumn{1}{c|}{2.950}                                 & 2,508,800                                                           \\ \hline
                                                                           & 5       & $32\times32\times256$  & $4\times4\times256\times128$                                                           & \multicolumn{1}{l|}{0.172694}                                                     & \multicolumn{1}{l|}{0.042322}                                                  & \multicolumn{1}{l|}{12.197}                                                       & \multicolumn{1}{c|}{2.866}                                & 4,596,736                                                           \\ \hline
                                                                           & 6       & $64\times64\times128$  & $4\times4\times128\times64$                                                            & \multicolumn{1}{l|}{0.175486}                                                     & \multicolumn{1}{l|}{0.041105}                                                  & \multicolumn{1}{c|}{11.745}                                                       &  \ 2.774                                                     & 8,786,432                                                           \\ \hline
                                                                           & 7       & $128\times128\times64$ & $4\times4\times64\times64$                                                             & \multicolumn{1}{l|}{0.349605}                                                     & \multicolumn{1}{l|}{0.082192}                                                  & \multicolumn{1}{c|}{22.398}                                                       & \ 5.233                                                     & 17,172,736                                                          \\ \hline
                                                                           &         &            & Total                                                                 & \multicolumn{1}{l|}{1.236147}                                                     & \multicolumn{1}{l|}{0.302276}                                                  & \multicolumn{1}{l|}{93.2194}                                                      & 20.34                                                     & \multicolumn{1}{c|}{}                                               \\ \hline
                                                                           &         &            & \begin{tabular}[c]{@{}l@{}}Total Speedup/\\ Memory Saved\end{tabular} & \multicolumn{1}{l|}{\textbf{}}                                                    & \multicolumn{1}{l|}{4.089464}                                                  & \multicolumn{1}{l|}{}                                                             & 4.583                                                     & \multicolumn{1}{c|}{35,534,592}                                     \\ \hline
\end{tabular}
\caption{Speedup for GPU and CPU versions and memory savings obtained from transpose convolution layers for popular GAN models}
\label{tab_7}
\end{small}

\end{table*}
\begin{table*}[!htbp]
\centering
\begin{small}
\begin{tabular}{|l|l|l|l|l|l|}
\hline
Model                                                                           & Layer \# & \begin{tabular}[c]{@{}l@{}}\# of \\ multiplications \\ (original)\end{tabular} & \begin{tabular}[c]{@{}l@{}}\# of \\ multiplications   \\ (proposed)\end{tabular} & \begin{tabular}[c]{@{}l@{}}\# of\\  additions   \\ (original)\end{tabular} & \begin{tabular}[c]{@{}l@{}}\# of \\ additions\\  (proposed)\end{tabular} \\ \hline
\begin{tabular}[c]{@{}l@{}}DCGAN\\ /DISCOGAN\end{tabular}                       & 2     & 536,870,912                                                                          & 134,217,728                                                                            & 536,838,144                                                                      & 134,184,960                                                                    \\ \hline
                                                                                & 3     & 536,870,912                                                                          & 134,217,728                                                                            & 536,805,376                                                                      & 134,152,192                                                                    \\ \hline
                                                                                & 4     & 536,870,912                                                                          & 134,217,728                                                                            & 536,739,840                                                                      & 134,086,656                                                                    \\ \hline
                                                                                & 5     & 25,165,824                                                                           & 6,291,456                                                                              & 25,153,536                                                                       & 6,279,168                                                                      \\ \hline
Total                                                                           &       & 1,635,778,560                                                                         & 408,944,640                                                                            & 1,635,536,896                                                                     & 408,702,976                                                                    \\ \hline
\begin{tabular}[c]{@{}l@{}}Total \\ \# of reductions \\ in operations\end{tabular} &       &                                                                                    & 1,226,833,920                                                                           &                                                                                & 1,226,833,920                                                                   \\ \hline
Art-GAN                                                                         & 2     & 134,217,728                                                                          & 33,554,432                                                                             & 134,201,344                                                                      & 33,538,048                                                                     \\ \hline
                                                                                & 3     & 134,217,728                                                                          & 33,554,432                                                                             & 134,184,960                                                                      & 33,521,664                                                                     \\ \hline
                                                                                & 4     & 268,435,456                                                                          & 67,108,864                                                                             & 268,304,384                                                                      & 66,977,792                                                                     \\ \hline
                                                                                & 6     & 25,165,824                                                                           & 6,2914,56                                                                              & 25,153,536                                                                       & 6,279,168                                                                      \\ \hline
Total                                                                           &       & 562,036,736                                                                          & 140,509,184                                                                            & 561,844,224                                                                      & 140,316,672                                                                    \\ \hline
\begin{tabular}[c]{@{}l@{}}Total \\ \# of reductions \\ in operations\end{tabular} &       &                                                                                    & 421,527,552                                                                            &                                                                                & 421,527,552                                                                    \\ \hline
 GP-GAN                                                                          & 2     & 134,217,728                                                                          & 33,554,432                                                                             & 134,201,344                                                                      & 33,538,048                                                                     \\ \hline
                                                                                & 3     & 134,217,728                                                                          & 33,554,432                                                                             & 134,184,960                                                                      & 33,521,664                                                                     \\ \hline
                                                                                & 4     & 134,217,728                                                                          & 33,554,432                                                                             & 134,152,192                                                                      & 33,488,896                                                                     \\ \hline
                                                                                & 5     & 12,582,912                                                                           & 3,145,728                                                                              & 12,570,624                                                                       & 3,133,440                                                                      \\ \hline
Total                                                                           &       & 415,236,096                                                                          & 103,809,024                                                                            & 415,109,120                                                                      & 103,412,048                                                                    \\ \hline
\begin{tabular}[c]{@{}l@{}}Total \\ \# of reductions\\  in operations\end{tabular} &       &                                                                                    & 311,427,072                                                                            &                                                                                & 311,697,072                                                                    \\ \hline
 EB-GAN                                                                          & 2     & 2,147,483,648                                                                         & 536,870,912                                                                            & 2,147,418,112                                                                     & 536,805,376                                                                    \\ \hline
                                                                                & 3     & 2,147,483,648                                                                         & 536,870,912                                                                            & 2,147,352,576                                                                     & 536,739,840                                                                    \\ \hline
                                                                                & 4     & 2,147,483,648                                                                         & 536,870,912                                                                            & 2,147,221,504                                                                     & 536,608,768                                                                    \\ \hline
                                                                                & 5     & 2,147,483,648                                                                         & 536,870,912                                                                            & 2,146,959,360                                                                     & 536,346,624                                                                    \\ \hline
                                                                                & 6     & 2,147,483,648                                                                         & 536,8709,12                                                                            & 2,146,435,072                                                                     & 535,822,336                                                                    \\ \hline
                                                                                & 7     & 4,294,967,296                                                                         & 1,073,741,824                                                                           & 4,290,772,992                                                                     & 1,069,547,520                                                                   \\ \hline
Total                                                                           &       & 15,032,385,536                                                                        & 3,758,096,384                                                                           & 15,026,159,616                                                                    & 3,751,870,464                                                                   \\ \hline
\begin{tabular}[c]{@{}l@{}}Total \\ \# of reductions\\  in operations\end{tabular} &       &                                                                                    & 11,274,289,152                                                                          &                                                                                & 11,274,289,152                                                                   \\ \hline
\end{tabular}
\caption{Number of floating point operations like multiplications and additions required for the conventional and proposed methods}
\label{tab_22}
\end{small}
\end{table*} 
\begin{table*}[!htbp]
\centering
\begin{small}
\begin{tabular}{|lllllll|}
\hline
\multicolumn{1}{|l|}{}                                                         & \multicolumn{3}{l|}{45nm technology}                                                                                                                                                                                                        & \multicolumn{3}{l|}{14nm technology}                                                                                                                                                                                   \\ \hline
\multicolumn{1}{|l|}{Design}                                                   & \multicolumn{1}{l|}{\begin{tabular}[c]{@{}l@{}}Delay (ns)\end{tabular}} & \multicolumn{1}{l|}{\begin{tabular}[c]{@{}l@{}}Area\\ (cell  units)\end{tabular}} & \multicolumn{1}{l|}{\begin{tabular}[c]{@{}l@{}}Power (mW)\end{tabular}} & \multicolumn{1}{l|}{\begin{tabular}[c]{@{}l@{}}Delay (ns)\end{tabular}} & \multicolumn{1}{l|}{\begin{tabular}[c]{@{}l@{}}Area\\ (cell  units)\end{tabular}} & \begin{tabular}[c]{@{}l@{}}Power (mW)\end{tabular} \\ \hline
\multicolumn{7}{|c|}{$3\times3$ kernel}                                                                                                                                                                                                                                                                                                                                                                                                                                                                                                                      \\ \hline
\multicolumn{1}{|l|}{\begin{tabular}[c]{@{}l@{}}Conventional\\ / 1 output\end{tabular}}  & \multicolumn{1}{l|}{1.53}                                                 & \multicolumn{1}{l|}{29413.37}                                                       & \multicolumn{1}{l|}{19.23}                                                & \multicolumn{1}{l|}{0.49}                                                 & \multicolumn{1}{l|}{3105.55}                                                        & 2.93                                                 \\ \hline
\multicolumn{1}{|l|}{\begin{tabular}[c]{@{}l@{}}Proposed\\ / 4 outputs\end{tabular}} & \multicolumn{1}{l|}{1.31}                                                 & \multicolumn{1}{l|}{29019.63}                                                       & \multicolumn{1}{l|}{19.91}                                                & \multicolumn{1}{l|}{0.44}                                                 & \multicolumn{1}{l|}{3070.52}                                                        & 2.96                                                 \\ \hline
\multicolumn{7}{|c|}{$4\times4$ kernel}                                                                                                                                                                                                                                                                                                                                                                                                                                                                                                                      \\ \hline
\multicolumn{1}{|l|}{\begin{tabular}[c]{@{}l@{}}Conventional\\/1 output\end{tabular}}  & \multicolumn{1}{l|}{1.66}                                                 & \multicolumn{1}{l|}{54174.12}                                                       & \multicolumn{1}{l|}{31.90}                                                & \multicolumn{1}{l|}{0.52}                                                 & \multicolumn{1}{l|}{5835.89}                                                        & 5.19                                                 \\ \hline
\multicolumn{1}{|l|}{\begin{tabular}[c]{@{}l@{}}Proposed\\/4 outputs\end{tabular}} & \multicolumn{1}{l|}{1.35}                                                 & \multicolumn{1}{l|}{51217.06}                                                       & \multicolumn{1}{l|}{37.57}                                                & \multicolumn{1}{l|}{0.44}                                                 & \multicolumn{1}{l|}{5645.68}                                                        & 5.70                                                 \\ \hline
\multicolumn{7}{|c|}{$5\times5$ kernel}                                                                                                                                                                                                                                                                                                                                                                                                                                                                                                                      \\ \hline
\multicolumn{1}{|l|}{\begin{tabular}[c]{@{}l@{}}Conventional\\/1 output\end{tabular}}  & \multicolumn{1}{l|}{1.77}                                                 & \multicolumn{1}{l|}{78509.66}                                                       & \multicolumn{1}{l|}{46.48}                                                & \multicolumn{1}{l|}{0.54}                                                 & \multicolumn{1}{l|}{8966.66}                                                        & 7.77                                                 \\ \hline
\multicolumn{1}{|l|}{\begin{tabular}[c]{@{}l@{}}Proposed\\/4 outputs\end{tabular}} & \multicolumn{1}{l|}{1.52}                                                 & \multicolumn{1}{l|}{71270.24}                                                       & \multicolumn{1}{l|}{56.38}                                                & \multicolumn{1}{l|}{0.49}                                                 & \multicolumn{1}{l|}{8549.79}                                                        & 8.31                                                 \\ \hline
\end{tabular}
\caption{Synthesis results of the conventional and proposed methods using 45nm and 14nm technologies with three different integer kernels} 
\label{t_3}
\end{small}
\end{table*}
\begin{table*}[!htbp]
\centering
\begin{small}
\begin{tabular}{|l|l|l|l|l|l|}
\hline
Original model    & Shape                                                                 & Modified model & Shape                                                               & Proposed model                                                                & Shape                                                                                \\ \hline
Input layer             & 28×28×1                                                           & Input layer           & 28×28×1                                                         & Input layer                                                                          & 28×28×1                                                                          \\ \hline
------------- & ---------                                                   & Upsampling layer        & 55×55×1                                                         & \multirow{2}{*}{\begin{tabular}[c]{@{}l@{}}Proposed optimized \\ layer\end{tabular}} & \multirow{2}{*}{\begin{tabular}[c]{@{}l@{}}51 × 51 × 8 \\ (5×5\\ kernel)\end{tabular}} \\ \cline{1-4}
CONV layer     & \begin{tabular}[c]{@{}l@{}}24×24×8 \\ (5×5\\ kernel)\end{tabular} & CONV layer   & \begin{tabular}[c]{@{}l@{}}51×51×8 \\ (5×5\\ kernel)\end{tabular} &                                                                                      &                                                                                      \\ \hline
ReLU   & 24×24×8                                                           & ReLU & 51×51×8                                                         & ReLU                                                              & 51×51×8                                                                          \\ \hline
Max pooling             & 12×12×8                                                           & Max pooling           & 26×26×8                                                         & Max pooling                                                                          & 26×26×8                                                                          \\ \hline
FC layer   & 10                                                                    & FC layer & 10                                                                  & FC layer                                                                & 10                                                                                   \\ \hline
\end{tabular}
\caption{Simple Deep Learning model configuration along with the total number of neurons for the MNIST dataset}
 \label{tab_5}
\end{small}

\end{table*}

% Please add the following required packages to your document preamble:
% \usepackage{multirow}

\subsection{Datasets and the evaluation procedure}  
We considered the flower dataset from the Kaggle website  \cite{mamaev}, MSCOCO 2017 \cite{cocodataset}, and PASCAL VOC 2012 \cite{pascal} datasets to compare the computation times and memory savings for the conventional and proposed optimized approaches for transpose convolution operation. The flower dataset contains five subgroups of classes: sunflower, dandelion, daisy, rose, and tulip. The total number of images in this dataset is 4,323. The sunflower class contains 734; the tulip class includes 984; the daisy class contains 769; the rose class contains 784; and the dandelion class contains 1,052 color images. We considered only 10\% of the available images, 11,828, from the MSCOCO 2017 dataset for the experimental analysis. Also, for the PASCAL 2017 dataset, we used both classification and segmentation datasets. The classification dataset contains 17,125 images, whereas the segmentation dataset contains 2,913 images of various sizes. For standard evaluation, all the images from the selected datasets are transformed into a standard format of $224\times224\times3$. We applied transpose convolution to the images and assessed the computation time using the conventional and the proposed methods. The programming languages used here were C++ and CUDA C for the CPU and GPU, respectively. The computation time and memory requirements are considered for evaluating the benefits of the proposed approach with the conventional implementation. \vspace{1mm}

\subsection{Analysis of computation time and memory savings}

Compared to the conventional approach, speedup and memory savings from the proposed optimization process with the selected datasets can be seen in Tables \ref{tab:1} and \ref{tab:2}. We used the Intel Xeon CPU and Nvidia GeForce RTX 2070 GPU for experimental analysis with GCC 9.4 and CUDA 10.2 versions, respectively. We varied the kernel size of $5\times5$, $4\times4$, and $3\times3$ to apply the transpose convolution operation for the input dimension of $224\times224\times3$. We reported the flower dataset's computation time and memory savings obtained from both approaches. The results showed that the sub-classes of the flower dataset reached $3.4\times$ ($3.7\times$) speedup on average for GPU (CPU), with memory savings above 11,824,304  bytes based on the kernel size. Similarly, the average speedup of $3.4\times$ ($3.8\times$) for GPU(CPU) was achieved for the MSCOCO 2017 and PASCAL VOC 2012 datasets. Since all the input images for these datasets are preprocessed into the exact size of $224\times 224\times3$, the memory savings still holds the same for these datasets from Table \ref{tab:1}. The speedup is significantly improved with the increase in the kernel sizes for all three datasets, with the corresponding memory savings. However, the even order kernel showed more memory savings because it didn't produce offset elements during computation.

\subsection{Ablation study} \vspace{-3mm}
The computation time, memory savings, and computation load for the transpose convolution layers commonly used in the popular GAN architectures \cite{yazdanbakhsh2018ganax} are reported in Tables \ref{tab_7} and \ref{tab_22}. The forward propagation phase for the layers is only considered by taking only one input sample during experimental analysis. In the DC-GAN/DiscoGAN, the average speedup of $3.9\times$ ($4.34\times$) was achieved for GPU (CPU) from the proposed approach with the overall memory savings of 4,787,712 bytes from the transpose convolution layers. Similarly, Art-GAN and GP-GAN got an average speedup of $2.95\times$ ($4.2\times$) for GPU (CPU). EB-GAN model showed the highest speedup of $4.08\times$ ($4.583\times$) because of more computation load needed for the transpose convolution layers in the model. We obtained limited GPU speedup for Art-GAN and GP-GAN models since the number of floating point operations like multiplications and additions is relatively less than in other models, which results in lower memory transfers. Among all the analyzed models, EB-GAN showed the highest memory savings in bytes of 35,534,592 from all transpose convolution layers. Additionally, there will be considerable improvement in the speedup from the transpose convolution layers during the training process, especially from backward propagation.

% IMRAN: "kernel faster"? or "faster for the kernel 5x5". It's kinda confusing! 
% vijay: faster for both the kernels 3X3 and 5X5 changed

% Please add the following required packages to your document preamble:
% \usepackage{multirow}
% Please add the following required packages to your document preamble:
% \usepackage{multirow}

\subsection{Hardware implementation}

The functional unit for the transpose convolution operation is implemented using the Verilog language to understand the hardware characteristics for the conventional and proposed optimization methods, as depicted in Table \ref{t_3}. Here, Synopsys DC Compiler with 45nm and 14nm technology is used to analyze the original and proposed methods' performance using integer kernels of 32 bits with an input size of 8 bits. Results indicate that the proposed model requires more power but less delay and area than the conventional implementation. However, the power consumption for the proposed method is higher because it writes four output values instead of one, compared to the conventional implementation.

\subsection{Evaluation using a simple neural network model}

We evaluated the training time using a simple convolutional neural network model for practical application in deep learning to illustrate the advantage of the proposed optimization. The model design having one convolutional layer trained on the MNIST dataset \cite{LeCun} is considered for the analysis, and the model's structure can be seen in Table. \ref{tab_5}. It consists of an input layer with a shape of $28 \times 28 \times 1$ followed by a convolutional layer (CONV layer) with a Rectified Linear Unit (ReLU) as an activation function and a max-pooling layer. Finally, a fully connected layer (FC layer) is added with ten neurons, as there are ten classes of MNIST images. Later, the convolutional layer is replaced with conventional and proposed transpose convolution layers to compare the training time for both models. The model was trained using Intel dual-core processor with all the layers implemented using C++. The training time is taken for the model when 100,000 epochs with a minibatch size of 1 for comparing the two models. The training time obtained for the original model was 1,100 seconds, whereas the proposed model took only 501 seconds. Results showed that our proposed optimized algorithm performed $2.2 \times$ faster than the conventional approach.

\section{Conclusion and future work} \label{sec6}
This manuscript proposed a novel optimization technique for transpose convolution operations using the kernel segregation mechanism. And it obtained an average speedup of $3.7 \times$ ($3.4\times$) on computation time for the CPU (GPU) compared to the naive transpose convolution implementation with notable memory savings. Furthermore, the optimized technique is applied to the simple deep learning model, which consists of a single transpose convolutional layer. The results indicated that the proposed method achieved $2.2\times$ faster than the conventional method. However, the proposed optimization method needs more power consumption than the traditional method as it writes four output values in the memory simultaneously instead of one. There is also a need to reduce power consumption for the proposed approach, which can be viewed as a future research direction. 
 \section*{Acknowledgement}
We sincerely thank Drs. Md Imran Hossen and Liqun Shan for their valuable time for the suggestions and for his dedicated help in organizing the manuscript effectively. We are also grateful to Dr.Nian-Feng Tzeng for his support. This work is supported in part by the US NSF under grants OIA-1946231 and CNS-2117785.

%Bibliography 
% \bibliographystyle{apalike}
% \bibliography{sample}
%\printbibliography
\bibliographystyle{apacite}
\footnotesize
\bibliography{sample}
\end{document}

%% file: hicss51-packages.tex
\usepackage[letterpaper]{geometry}
\usepackage{hicss51}
\usepackage{times}
\usepackage[none]{hyphenat}
\usepackage{url}
\usepackage{latexsym}
\usepackage{hyperref}
\usepackage{breakurl}
\hypersetup{
    colorlinks = true,
    linkcolor = red,
    anchorcolor = blue,
    citecolor = blue,
    filecolor = blue,
    urlcolor = blue
}
\usepackage[english]{babel}
\usepackage{apacite}
\usepackage{comment}
\usepackage{url}
\usepackage{latexsym}

\usepackage{graphicx}

\usepackage{array}
\usepackage{comment}
\usepackage{listings}
\usepackage{tabularx}
\usepackage{booktabs}
\usepackage{threeparttable}
\usepackage{multirow}
\usepackage{tabularx}
\usepackage{bigstrut}
\usepackage{amsmath}
\usepackage{amsfonts}
\usepackage{amssymb}
\usepackage{wasysym}
\usepackage{enumitem}
\usepackage{algorithm}
\usepackage[noend]{algorithmic}
\usepackage{xcolor}